\pdfoutput=1
\documentclass[11pt,a4paper]{article}
\usepackage[hyperref]{acl2019}
\usepackage{times}
\usepackage{latexsym}
\usepackage{pgfplots}

\usepackage{algorithm}
\usepackage[noend]{algpseudocode}

\aclfinalcopy 
\def\aclpaperid{30} 

\usepackage{amsmath}
\usepackage{amssymb}
\usepackage{bm}
\usepackage{todonotes}

\newcommand\h{\mathbf h}
\newcommand\z{\mathbf z}

\newcommand\E{\mathbb E}

\newcommand\DKL[2]{D^\text{\tiny KL}\!\left(#1||#2\right)}

\newcommand\R{\mathbb R}

\let\oldpm\pm
\renewcommand\pm{\ensuremath{\oldpm}}

\usepgfplotslibrary{groupplots}
\usetikzlibrary{matrix}

\definecolor{rosso}{RGB}{220,57,18}
\definecolor{giallo}{RGB}{255,153,0}
\definecolor{blu}{RGB}{102,140,217}
\definecolor{verde}{RGB}{16,150,24}
\definecolor{viola}{RGB}{153,0,153}
\definecolor{babyblue}{RGB}{0,129,255}
\definecolor{darkgreen}{RGB}{6,148,60}

\title{Generalization in Generation:  A closer look at Exposure Bias}

\author{
  Florian Schmidt \\
  Department of Computer Science\\
  ETH Z{\"u}rich\\
  \texttt{florian.schmidt@inf.ethz.ch}}
\date{}

\pgfplotsset{every y tick label/.append style={font=\tiny}}

\begin{document}
\maketitle
\begin{abstract}
\emph{Exposure bias} refers to the train-test discrepancy that seemingly arises when an autoregressive generative model uses only ground-truth contexts at training time but generated ones at test time. We separate the contributions of the model and the learning framework to clarify the debate on consequences and review proposed counter-measures. 

\vspace{-1.55mm}
In this light, we argue that generalization is the underlying property to address and propose unconditional generation as its fundamental benchmark. Finally, we combine latent variable modeling with a recent formulation of exploration in reinforcement learning to obtain a rigorous handling of true and generated contexts. Results on language modeling and variational sentence auto-encoding confirm the model's generalization capability.
\end{abstract}

\newcommand\p{{p_\theta}}
\newcommand\infp{{\tilde p}}
\newcommand\hiddenequal{\phantom{=\ }}
\newcommand\reward{R(w,w^\star)}
\newcommand\thetagrad{{\nabla_\theta}}

\section{Introduction}
\label{sec:introduction}

Autoregressive models span from $n$-gram models to recurrent neural networks to transformers and have formed the backbone of state-of-the-art machine learning models over the last decade on virtually any generative task in Natural Language Processing. Applications include machine translation \cite{bahdanauCB14,vaswaniSPUJGKP17}, summarization \cite{rushCW15, khandelwal2019}, dialogue \cite{serban2016building} and sentence compression \cite{filippova2015sentence}.

The training methodology of such models is rooted in the language modeling task, which is to predict a single word given a context of previous words. It has often been criticized that this setting is not suited for multi-step generation where -- at test time -- we are interested in generating words given a \emph{generated} context that was potentially not seen during training. The consequences of this train-test discrepancy are summarized as \emph{exposure bias}. Measures to mitigate the problem typically rely on replacing, masking or pertubing ground-truth contexts \cite{bengioVJS15, bowmanVVDJB15, norouziBCJSWS16, ranzato2015sequence}. Unfortunately, exposure bias has never been succesfully separated from general test-time log-likelihood assessment and minor improvements on the latter are used as the only signifier of reduced bias. Whenever explicit effects are investigated, no significant findings are made~\cite{he19exposure}.

In this work we argue that the standard training procedure, despite all criticism, is an immediate consequence of combining autoregressive modeling and maximum-likelihood training. As such, the paramount consideration for improving test-time performance is simply regularization for better generalization. In fact, many proposed measures against exposure bias can be seen as exactly that, yet with respect to an usually implicit metric that is not maximum-likelihood.

With this in mind, we discuss regularization for conditional and unconditional generation. We note that in conditional tasks, such as translation, it is usually sufficient to regularize the \emph{mapping} task -- here translation -- rather than the generative process itself. For unconditional generation, where tradeoffs between accuracy and coverage are key, generalization becomes much more tangible.\\

The debate on the right training procedure for autoregressive models has recently been amplified by the advent of \emph{latent} generative models \cite{rezendeMW14, kingma2013auto}.
Here, the practice of decoding with true contexts during training conflicts with the hope of obtaining a latent representation that encodes significant information about the sequence \cite{bowmanVVDJB15}. Interestingly, the ad hoc tricks to reduce the problem are similar to those proposed to address exposure bias in deterministic models.

Very recently, Tan et al.~\citeyearpar{Hu2017Towards} have presented a reinforcement learning formulation of exploration that allows following the intuition that  an autoregressive model should not only be trained on ground-truth contexts. We combine their framework with latent variable modeling and a reward function that leverages modern word-embeddings. The result is a single learning regime for unconditional generation in a deterministic setting (language modeling) and in a latent variable setting (variational sentence autoencoding). Empirical results show that our formulation allows for better generalization than existing methods proposed to address exposure bias. Even more, we find the resulting regularization to also improve generalization under log-likelihood. 

We conclude that it is worthwhile exploring reinforcement learning to elegantly extend maximum-likelihood learning where our desired notion of generalization cannot be expressed without violating the underlying principles. As a result, we hope to provide a more unified view on the training methodologies of autoregressive models and exposure bias in particular.

\begin{figure*}[t]
	\centering
	\includegraphics[scale=0.8]{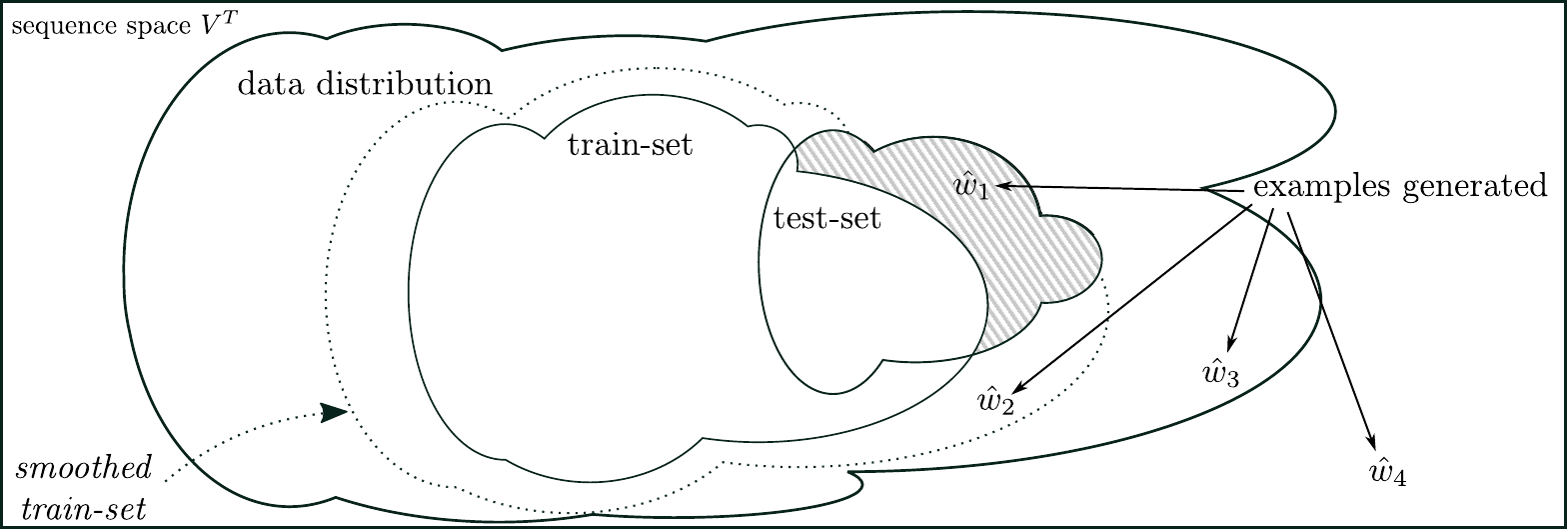}
	\caption{Generalization}
	\label{fig:generalization}
\end{figure*}

\section{Autoregressive Modeling}
\label{sec:intro:text-generation}
Modern text generation methods are rooted in models trained on the language modeling task. In essence, a \emph{language model} $p$ is trained to predict a word given its left-side context 
\begin{align}
	p(w_t|w_{1:t-1})\ .\label{eq:lmtask}
\end{align} 
With a trained language model at hand, a simple recurrent procedure allows to generate text of arbitrary length. Starting from an initial special symbol $\hat w_0$, we iterate $t=1\dots$ and alternate between sampling $\hat w_t\sim p(w_t|\hat w_{1:t-1})$ and appending $\hat w_t$ to the context $\hat w_{1:t-1}$. Models of this form are called \emph{autoregressive} as they condition new predictions on old predictions.

\paragraph{Neural Sequence Models}

Although a large corpus provides an abundance of word-context pairs to train on, the cardinality of the context space makes explicit estimates of \eqref{eq:lmtask} infeasible. Therefore, traditional $n$-gram language models rely on a truncated context and smoothing techniques to generalize well to unseen contexts. 

Neural language models lift the context restriction and instead use neural context representations. This can be a hidden state as found in recurrent neural networks (RNNs), i.e.\ an LSTM~\cite{hochreiterS1997} state, or a set of attention weights, as in a transformer architecture \cite{vaswaniSPUJGKP17}. While the considerations in this work apply to all autoregressive models, we focus on recurrent networks which encode the context in a fixed-sized continuous representation $\h(w_{1:t-1})$. In contrast to transformers, RNNs can be generalized easily to variational autoencoders with a single latent bottleneck \cite{bowmanVVDJB15}, a particularly interesting special case of generative models .

\subsection{Evaluation and Generalization}
\label{sec:generalization}

\paragraph{Conditional vs. Unconditional}\hspace{0.8cm}

\noindent Conditional generation tasks, such as translation or summarization, are attractive from an application perspective. However, for the purpose of studying exposure bias, we argue that unconditional generation is the task of choice for the following reasons.

First, exposure bias addresses conditioning on past words \emph{generated} which becomes less essential when words in a source sentence are available, in particular when attention is used.

Second, the difficulty of the underlying mapping task, say translation, is of no concern for the mechanics of generation. This casts sentence autoencoding as a less demanding, yet more economic task. 

Finally, generalization of conditional models is only studied with respect to the underlying mapping and not with respect to the conditional distribution itself. A test-set in translation usually does not contain a source sentence seen during training with a \emph{different} target\footnote{Some datasets do provide several targets for a single source. However, those are typically only used for BLEU computation, which is the standard test metric reported.}. Instead, it contains unseen source-target pairs that evaluate the generalization of the mapping. Even more, at test-time most conditional models resort to an arg-max decoding strategy. As a consequence, the entropy of the generative model is zero (given the source) and there is no generalization at all with respect to generation. For these reasons, we address unconditional generation and sentence auto-encoding for the rest of this work.

\paragraph{The big picture}

Let us briefly characterize output we should expect from a generative model with respect to generalization. Figure \ref{fig:generalization} shows an idealized two-dimensional dataspace of (fixed-length) sentences $w\in V^T$. We sketch the support of the unknown underlying generating distribution, the train set and the test set.%
\footnote{Here we do not discuss \emph{generalization error}, the discrepancy between empirical test error and expected test error. It should also be noted that cross-validation provides another complementary technique to more robust model estimation, which we omit to keep the picture simple.}
Let us look at some hypothetical examples $\hat w_1, \hat w_2,\hat w_3,\hat w_4$ generated from some well trained model. Samples like $\hat w_1$ certify that the model did not overfit to the training data as can be certified by test log-likelihood. In contrast, the remaining samples are indistinguishable under test log-likelihood in the sense that they identically decrease the metric (assuming equal model probability) even though $\hat w_2,\hat w_3$ have non-zero probability under the true data distribution. Consequently, we cannot identify $\hat w_4$ as a malformed example. Holtzman et al.~\citeyearpar{holtzman19} show that neural generative models -- despite their expressiveness -- put significant probability on clearly unreasonable repetitive phrases, such as \textit{I don’t know. I don’t know. I don’t know}.\footnote{They report that this also holds for non-grammatical repetitive phrase, which is what we would expect for $\hat w_4$.}

\paragraph{Evaluation under smoothed data distribution}

The most common approach to evaluating an unconditional probabilistic generative model is training and test log-likelihood. For a latent variable model, the exact log-likelihood \eqref{eq:marginal} is intractable and a lowerbound must be used instead. However, at this point it should be noted that one can always estimate the log-likelihood from an empirical distribution across output generated. That is, one generates a large set of sequences $\mathcal S$  and sets $\hat p(w)$ to the normalized count of $w$ in $\mathcal S$. However, the variance of this estimate is impractical for all but the smallest datasets. Also, even a large test-set cannot capture the flexibility and compositionality found in natural language.

With aforementioned shortcomings of test log-likelihood in mind, it is worthwhile discussing a recently proposed evaluation technique. Fedus et al.~\citeyearpar{fedusGD2018} propose to use $n$-gram statistics of the underlying data to asses generated output. For example, one can estimate an $n$-gram language model and report perplexity of the generated data under the $n$-gram model. Just as BLEU and ROUGE break the sequence reward assignment problem into smaller sub-problems, $n$-gram language models effectively \emph{smooth} the sequence likelihood assignment which is usually done with respect to the empirical data distribution. Under this metric, some sequences such as $\hat w_2$ which are close to sequences in the dataset at hand might receive positive probability.

This raises two questions. First, can we break sequence-level evaluation into local statistics by using modern word embeddings instead of $n$-grams (as BLEU does)? Second, can we incorporate these measures already during training to obtain better generative models. These considerations will be key when defining a reward function in Section \ref{sec:reward}.

\section{Teacher Forcing and Exposure Bias}
\label{sec:exposure-bias}
A concern often expressed in the context of autoregressive models is that the recursive sampling procedure for generation presented in Section \ref{sec:introduction} is never used at training time; hence the model cannot learn to digest its own predictions. The resulting potential train-test discrepancy is referred to as \emph{exposure bias} and is associated with compounding errors that arise when mistakes made early accumulate \cite{bengioVJS15, ranzato2015sequence, goyalLZZCB16, leblondAOL17}. In this context, \emph{teacher-forcing} refers to the fact that -- seen from the test-time perspective -- ground-truth contexts are substituted for model predictions. Although formally teacher forcing and exposure bias should be seen as cause (if any) and symptom, they are often used exchangeably.

As is sometimes but rarely mentioned, the presence of the ground-truth context is simply a consequence of maximum-likelihood training and the chain rule applied to \eqref{eq:lmtask} as in $p(w_{1:T})=\prod p(w_t|w_{1:t-1})$  \cite{dlbook}.  As such, it is out of question whether generated contexts should be used as long as log-likelihood is the sole criterion we care about. In this work we will furthermore argue the following:\\

\setlength{\leftskip}{0.4cm}

\noindent\textbf{Proposition 1}\ \textit{Exposure bias describes a lack of generalization with respect to an -- usually implicit and potentially task and domain dependent -- measure other than maximum-likelihood.}\\

\setlength{\leftskip}{0cm}

\noindent The fact that we are dealing with generalization is obvious, as one can train a model -- assuming sufficient capacity -- under the criticized methodology to match the training distribution.  Approaches that address exposure bias do not make the above notion of generalization explicit, but follow the intuition that training on other contexts than (only) ground-truth contexts should regularize the model and result in -- subjectively -- better results. Of course, these forms of regularization might still implement some form of log-likelihood regularization, hence improve log-likelihood generalization. Indeed, all of the following methods do report test log-likelihood improvements. 

\paragraph{Proposed methods against exposure bias}\hspace{0.5cm}

\noindent \textit{Scheduled sampling} \cite{bengioVJS15} proposed for conditional generation randomly mixes in predictions form the model, which violates the underlying learning framework \cite{Huszar2015HowT}. \textit{RAML}~\cite{norouziBCJSWS16} proposes to effectively perturbs the ground-truth context according to the exponentated payoff distribution implied by a reward function. Alternatively, adversarial approaches \cite{goyalLZZCB16} and learning-to-search  \cite{leblondAOL17} have been proposed.

Another line of work advocates \emph{globally} normalized sequence models to avoid biases induced by the locally normalized observations of standard autoregressive models \cite{wisemanR16, AndorAWSPGPC16, schmidt2019autoregressive}.

\paragraph{VAE Collapse}
In Section \ref{sec:latent-models} we will take a look at \emph{latent} generative models. In that context, the standard maximum-likelihood approach to autoregressive models has been criticized from a second perspective that is worth mentioning. Bowman et al.~\citeyearpar{bowmanVVDJB15} show empirically that autoregressive decoders $p(w|\z)$ do not rely on the latent code $\z$, but collapse to a language model as in \eqref{eq:lmtask}.

While some work argues that the problem is rooted in autoregressive decoders being ``too powerful"~\cite{shen2018}, the proposed measures often address the autoregressive training regime rather than the models \cite{bowmanVVDJB15} and, in fact, replace ground-truth contexts just as the above methods to mitigate exposure bias.

In addition, a whole body of work has discussed the implications of optimizing only a bound to the log-likelihood~\cite{alemi17} and the implications of re-weighting the information-theoretic quantities inside the bound \cite{Higgins2017betaVAELB, Rainforth2018TighterVB}.

\section{Latent Generation with ERPO}

We have discussed exposure bias and how it has been handled by either implicitly or explicitly leaving the maximum-likelihood framework. In this section, we present our reinforcement learning framework for unconditional sequence generation models. The generative story is the same as in a latent variable model:
\begin{enumerate}
	\item Sample a latent code $\z\sim\R^d$
	\item Sample a sequence from a code-conditioned policy $\p(w|\z)$.
\end{enumerate}
However, we will rely on reinforcement learning to train the decoder $p(w|\z$). Note that for a constant code $\z=\mathbf{0}$ we obtain a language model as a special case. Let us now briefly review latent sequential models.

\subsection{Latent sequential models}
\label{sec:latent-models}

Formally, a latent model of sequences $w=w_{1:T}$ is written as a marginal over latent codes
\begin{align}
	p(w)=\int p(w,\z)d\z=\int p(w|\z)p_0(\z)d\z\ .\label{eq:marginal}
\end{align}
The precise form of $p(w|\z)$ and whether $\z$ refers to a single factor or a sequence of factors $\z_{1:T}$ depends on the model of choice. 

The main motivation of enhancing $p$ with a latent factor is usually the hope to obtain a meaningful structure in the space of latent codes. How such a structure should be organized has been discussed in the \emph{disentanglement} literature in great detail, for example in Chen et al.~\citeyearpar{chen18}, Hu et al.~\citeyearpar{Hu2017Towards} or Tschannen et al.~\citeyearpar{tschannen2018}.

In our context, latent generative models are interesting for two reasons. First, explicitly introducing uncertainty inside the model is often motivated as a regularizing technique in Bayseian machine learning \cite{Murphy:2012} and has been applied extensively to latent sequence models~\cite{ziegler19, schmidt18, goyalSCKB17, bayer2014}. Second, as mentioned in Section \ref{sec:exposure-bias} (VAE collapse) conditioning on ground-truth contexts has been identified as detrimental to obtaining meaningful latent codes~\cite{bowmanVVDJB15} -- hence a methodology to training decoders that relaxes this requirement might be of value.

\paragraph{Training via Variational Inference}
Variational inference~\cite{zhang2018advances} allows to optimize a lower-bound instead of the intractable marginal likelihood and has become the standard methodology to training latent variable models. Introducing an inference model $q$ and applying Jensen's inequality to \eqref{eq:marginal}, we obtain
\begin{align}
	&\!\!\!\!\log p(w)=\E_{q(\z|w)}\!\left[\log\frac{p_0(\z)}{q(\z|w)}\! +\! \log P(w|\z)\right]\notag\\
	&\!\!\!\!\geq\DKL{q(\z|w)}{p_0(\z)} + \E_{q(\z|w)}\left[\log P(w|\z)\right]\!\!\!\label{eq:elbo}
\end{align}
Neural inference networks~\cite{rezendeMW14, kingma2013auto} have proven as effective amortized approximate inference models. 

Let us now discuss how reinforcement learning can help training our model.

\subsection{Generation as Reinforcement Learning}
Text generation can easily be formulated as a reinforcement learning (RL) problem if words are taken as actions~\cite{bahdanauBXGLPCB16}. Formally, $\p$ is a parameterized policy that factorizes autoregressively $\p(w)=\prod\p(w_t|\h(w_{1:t-1}))$ and $\h$ is is a deterministic mapping from past predictions to a continuous state, typically a recurrent neural network (RNN). The goal is then to find policy parameters $\theta$ that maximize the expected reward 
\begin{align}
	J(\theta)=\E_{\p(w)}[\reward]	\label{eq:reward-maximization}
\end{align}
where $\reward$ is a task-specific, not necessarily differentiable metric.
\paragraph{Policy gradient optimization} The REINFORCE \cite{Williams1992} training algorithm is a common strategy to optimize \eqref{eq:reward-maximization} using a gradient estimate via the log-derivative
\begin{align}
	\thetagrad J(\theta)=\E_{\p(w)}[\reward\log\p(w)]\label{eq:estimator}
\end{align}
Since samples from the policy $\hat w\sim\p$ often yield low or zeros reward, the  estimator \eqref{eq:estimator} is known for its notorious variance and much of the literature is focused on reducing this variance via baselines or control-derivative~\cite{rennieMMRG16}.

\subsection{Reinforcement Learning as Inference}
Recently, a new family of policy gradient methods has been proposed that draws inspiration from inference problems in probablistic models. The underlying idea is to pull the reward in \eqref{eq:estimator} into a new \emph{implicit} distribution $\infp$ that allows to draw samples $\hat w$ with much lower variance as it is informed about reward.

We follow Tan et al.~\citeyearpar{Hu2017Towards} who optimize an entropy-regularized version of \eqref{eq:reward-maximization}, a common strategy to foster exploration. They cast the reinforcement learning problem as
\begin{align}
	J(\theta,\infp) &= \E_\infp[R(w,w^\star)]\notag\\ 
				&\hiddenequal+ \alpha \DKL{\infp(w)}{\p(w)} \notag\\
				&\hiddenequal+ \beta H(\infp)\label{eq:er-reward}
\end{align}	 
where $\alpha,\beta$ are hyper-parameters and $\infp$ is the new non-parametric, \emph{variational} distribution\footnote{In \cite{tan2018} $\infp$ is written as $q$, which resembles variational distributions in approximate Bayesian inference. However, here $\infp$ is not defined over variables but datapoints.} across sequences. They show that \eqref{eq:er-reward} can be optimized using the following EM updates
\begin{align}
	\!\!\!\!\!\!\!\text{E-step:}\,\infp^{n+1}&\!\!\propto\exp\left(\frac{\alpha\p^n(w) + R(w,w^\star)}{\alpha + \beta}\right)\!\!\!\!\!\!\label{eq:e-step}\\
	\!\!\!\!\!\text{M-step:}\,\theta^{n+1}&\!=\!\arg\max_\theta\E_{\infp^{n+1}}[\log \p(w)]
\end{align}
As Tan et al.~\citeyear{tan2018} have shown, for $\alpha\rightarrow0$, $\beta=1$ and a specific reward, the framework recovers maximum-likelihood training.\footnote{Refer to their work for more special cases, including MIXER \cite{ranzato2015sequence}} It is explicitly not our goal to claim text generation with end-to-end reinforcement learning but to show that it is beneficial to operate in an RL regime relatively close to maximum-likelihood.

\subsection{Optimization with Variational Inference}	
In conditional generation, a policy is conditioned on a source sentence, which guides generation towards sequences that obtain significant reward. Often, several epochs of MLE pretraining~\cite{rennieMMRG16, bahdanauBXGLPCB16} are necessary to make this guidance effective.

In our unconditional setting, where a source is not available, we employ the latent code $\z$ to provide guidance. We cast the policy $\p$ as a code-conditioned policy $\p(w|\z)$ which is trained to maximize a marginal version of the reward \eqref{eq:er-reward}:
\begin{align}
	J(\theta)=\E_{p_0(\z)}\E_{\p(w|\z)}[\reward]]\ .\label{eq:expected-reward}
\end{align}
Similar formulations of expected reward have recently been proposed as \emph{goal-conditioned} policies~\cite{ghosh18}. However, here it is our explicit goal to also learn the representation of the goal, our latent code. We follow Equation \eqref{eq:elbo} and optimize a lower-bound instead of the intractable marginalization \eqref{eq:expected-reward}. Following \cite{bowman15snli, fraccaro2016SPW} we use a deep RNN inference network for $q$ to optimize the bound. The reparametrization-trick~\cite{kingma2013auto} allows us to compute gradients with respect to $q$. Algorithm \ref{algorithm} shows the outline of the training procedure.

\begin{algorithm}[ht]           
\caption{Latent ERPO Training}     
\label{alg1}
\begin{algorithmic}  
\For \; $w^\star\in\textsc{dataset}$
	\State Sample a latent code \hspace{7.5mm}$\z\sim q(\z|w^\star)$
	\State Sample a datapoint \hspace{8mm} $\tilde w\sim\infp(w|\z)$
	\State Perform a gradient step  \hspace{2.5mm}$\nabla_\theta\log\p(\tilde w|\z)$
\EndFor
\end{algorithmic}
\label{algorithm}
\end{algorithm}

Note that exploration (sampling $\tilde w$) and the gradient step are both conditioned on the latent code, hence stochasticity due to sampling a single $\z$ is coupled in both. Also, no gradient needs to be propagated into $\infp$. 

So far, we have not discussed how to efficiently sample from the implicit distribution $\infp$. In the remainder of this section we present our reward function and discuss implications on the tractability of sampling.

\subsection{Reward} 
\label{sec:reward}

Defining a meaningful reward function is central to the success of reinforcement learning. The usual RL forumlations in NLP require a measure of sentence-sentence similarity as reward. Common choices include BLEU~\cite{bleu}, ROUGE~\cite{rouge}, CIDEr~\cite{banerjee-lavie-2005-meteor} or SPICE~\cite{andersonFJG16}. These are essentially $n$-gram metrics, partly augmented with synonym resolution or re-weighting schemes. 

Word-movers distance (WMD)~\cite{Kusner2015} provides an interesting alternative based on the optimal-transport problem. In essence, WMD computes the minimum accumulated distance that the word vectors of one sentence need to ``travel" to coincide with the word vectors of the other sentence. In contrast to $n$-gram metrics, WMD can leverage powerful neural word representations. Unfortunately, the complexity of computing WMD is roughly $\mathcal O(T^3\log T)$.

\begin{figure*}[ht]

\centering
\begin{tikzpicture}
\hspace{-0.35cm}
\begin{groupplot}[
    group style = {group size = 2 by 1, horizontal sep=1.cm},
    xlabel={Training Time},
    height = 6cm,
    width=8.6	cm,
    ylabel style={yshift=-0.3cm},
    xlabel style={yshift=0.2cm},
    legend cell align={left},
    legend style={font=\scriptsize, row sep=-2.4pt},
    xmin=20000,
    xmax=180000,
    xticklabels={,,},
    restrict x to domain=20000:180000,
    scaled x ticks=base 10:0,
    xticklabel style={/pgf/number format/fixed},
    ylabel style={font=\small,yshift=-0.35cm},
    xlabel style={font=\small,yshift=0.3cm},
]

\pgfplotstableread{latent_train.data}{\datatable}
\nextgroupplot[ylabel=Train NLL,ymin=47.5, ymax=54]

\addplot [babyblue] table [x={x}, y={latent-ss-p0.99}] {\datatable};
\label{plot:latent-ss-p0.99}
\addplot [babyblue, dashed] table [x={x}, y={latent-ss-p0.98}] {\datatable};
\label{plot:latent-ss-p0.98}
\addplot [babyblue, dotted] table [x={x}, y={latent-ss-p0.95}] {\datatable};
\label{plot:latent-ss-p0.95}
\addplot [babyblue, loosely dotted] table [x={x}, y={latent-ss-p0.9}] {\datatable};
\label{plot:latent-ss-p0.9}
\addplot [black] table [x={x}, y={latent-vae}] {\datatable};
\label{plot:latent-vae}
\addplot [gray] table [x={x}, y={latent-wordrop-0.99}] {\datatable};
\label{plot:latent-wordrop-0.99}
\addplot [giallo, thick] table [x={x}, y={latent-alpha0.01-beta0.1-gamma1.5}] {\datatable};
\label{plot:latent-alpha0.01-beta0.1-gamma1.5}
\addplot [verde] table [x={x}, y={latent-raml-beta0.065}] {\datatable};
\label{plot:latent-raml-beta0.065}           
\addplot [rosso, thick] table [x={x}, y={latent-alpha0.015-beta0.1-gamma1.4}] {\datatable};
\label{plot:latent-alpha0.015-beta0.1-gamma1.4}

\addplot [viola, thick,mark repeat=1,mark phase=2,mark=x,mark size=2pt] table [x={x}, y={lm-alpha0.1-beta0.05-gamma1.8}] {\datatable};
                  \label{plot:lm-alpha0.1-beta0.05-gamma1.8}
\addplot [babyblue,mark repeat=1,mark phase=2,mark=x,mark size=2pt] table [x={x}, y={lm-ss-p0.99}] {\datatable};
                  \label{plot:lm-ss-p0.99}
\addplot [black,mark repeat=1,mark phase=2,mark=x,mark size=2pt] table [x={x}, y={lm}] {\datatable};
                  \label{plot:lmx}
\addplot [verde,mark repeat=1,mark phase=2,mark=x,mark size=2pt] table [x={x}, y={lm-raml-beta0.06}] {\datatable};
                  \label{plot:lm-raml-beta0.06}

\coordinate (top) at (rel axis cs:0,1);
\nextgroupplot[ylabel=Test NLL, ylabel style={yshift=0.15cm},ymin=52.5, ymax=54]
\pgfplotstableread{latent_test.data}{\datatable}

\addplot [babyblue] table [x={x}, y={latent-ss-p0.99}] {\datatable};
\label{plot:latent-ss-p0.99}
\addplot [babyblue, dashed] table [x={x}, y={latent-ss-p0.98}] {\datatable};
\label{plot:latent-ss-p0.98}
\addplot [babyblue, dotted] table [x={x}, y={latent-ss-p0.95}] {\datatable};
\label{plot:latent-ss-p0.95}
\addplot [babyblue, loosely dotted] table [x={x}, y={latent-ss-p0.9}] {\datatable};
\label{plot:latent-ss-p0.9}
\addplot [black] table [x={x}, y={latent-vae}] {\datatable};
\label{plot:latent-vae}
\addplot [gray] table [x={x}, y={latent-wordrop-0.99}] {\datatable};
\label{plot:latent-wordrop-0.99}
\addplot [giallo, thick] table [x={x}, y={latent-alpha0.01-beta0.1-gamma1.5}] {\datatable};
\label{plot:latent-alpha0.01-beta0.1-gamma1.5}
\addplot [verde] table [x={x}, y={latent-raml-beta0.065}] {\datatable};
\label{plot:latent-raml-beta0.065}           
\addplot [rosso, thick] table [x={x}, y={latent-alpha0.015-beta0.1-gamma1.4}] {\datatable};
\label{plot:latent-alpha0.015-beta0.1-gamma1.4}

\addplot [viola, thick,mark repeat=1,mark phase=2,mark=x,mark size=2pt] table [x={x}, y={lm-alpha0.1-beta0.05-gamma1.8}] {\datatable};
                  \label{plot:lm-alpha0.1-beta0.05-gamma1.8}
\addplot [babyblue,mark repeat=1,mark phase=2,mark=x,mark size=2pt] table [x={x}, y={lm-ss-p0.99}] {\datatable};
                  \label{plot:lm-ss-p0.99}
\addplot [black,mark repeat=1,mark phase=2,mark=x,mark size=2pt] table [x={x}, y={lm}] {\datatable};
                  \label{plot:lmx}
\addplot [verde,mark repeat=1,mark phase=2,mark=x,mark size=2pt] table [x={x}, y={lm-raml-beta0.06}] {\datatable};
                  \label{plot:lm-raml-beta0.06}       

\coordinate (bot) at (rel axis cs:1,0);
\end{groupplot}

 \path (top-|current bounding box.west)-- 
          node[anchor=south,rotate=90] {} 
          (bot-|current bounding box.west);
\path (top|-current bounding box.north)--
      coordinate(legendpos)
      (bot|-current bounding box.north);
\matrix[
    matrix of nodes,
    anchor=south,
    draw,
    inner sep=0.1em,
    column sep =0.3em,
    every node/.style={anchor=base west},
    draw
  ]at([yshift=0.0ex]legendpos)
  {
    \ref{plot:latent-alpha0.01-beta0.1-gamma1.5} &\small\sc Ours &[20pt]
    \ref{plot:latent-alpha0.015-beta0.1-gamma1.4}& \small\sc Ours-B &[20pt]
    \ref{plot:lm-alpha0.1-beta0.05-gamma1.8} &  \small\sc Ours-det &[2pt]\\
    \ref{plot:latent-ss-p0.99}& \small\sc SS-0.99 &[20pt]
   	\ref{plot:latent-ss-p0.98}& \small\sc SS-0.98 &[20pt]
    \ref{plot:latent-ss-p0.95}& \small\sc SS-0.95 &[20pt]
    \ref{plot:latent-ss-p0.9}& \small\sc SS-0.90 &[20pt]
    \ref{plot:lm-ss-p0.99}&  \small\sc ss-0.99-det &[2pt]\\
    \ref{plot:latent-raml-beta0.065}& \small\sc RAML &[20pt]
    \ref{plot:lm-raml-beta0.06}&  \small\sc RAML-det &[2pt]
	\ref{plot:latent-vae}& \small\sc VAE &[20pt]
    \ref{plot:latent-wordrop-0.99}& \small\sc wdrop-0.99 &[20pt]
    \ref{plot:lmx}&  \small\sc LM-det &[2pt]\\
    };

\end{tikzpicture}
\caption{Generalization performance in terms of sequence NLL across latent and deterministic methods}
\label{plot:latent}
\end{figure*}
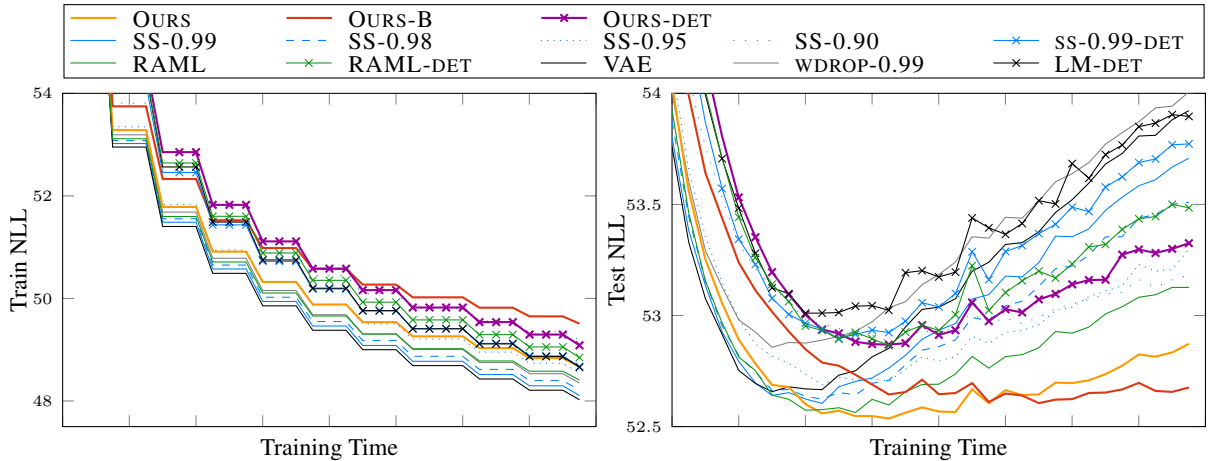

\subsection{A Reward for Tractable Sampling}
Tan et al.~\citeyearpar{tan2018} show that thanks to the factorization of $\p$ the globally-normalized inference distribution $\infp$ in \eqref{eq:e-step} can be written as a locally-normalized distribution at the word-level
\begin{align}
\!\!\!\!\infp(w_t|&w_{1:t-1})\!\propto\notag\\
&\hspace{2mm}\exp\left(\frac{\alpha\p(w_t|w_{1:t-1})\! +\! R_t(w,w^\star)}{\alpha + \beta}\right)\label{eq:word-sampling}
\end{align}
when the reward is written as incremental reward $R_t$  defined via $R_t(w,w^\star) = R(w_{1:t},w^\star) - R(w_{1:t-1},w^\star)$. Sampling form \eqref{eq:word-sampling} is still hard, if $R_t$ hides dynamic programming routines or other complex time-dependencies. With this in mind, we choose a particularly simple reward 
\begin{align}
	R(w,w^\star)=\sum_{t=1}^T\phi(w_t)^\top\phi(w_t^\star)\label{eq:reward}
\end{align}
where $\phi$ is a lookup into a length-normalized pre-trained but fixed word2vec \cite{mikolov13} embedding. This casts our reward as an efficient, yet drastic approximation to WMD, which assumes identical length and one-to-one word correspondences. Putting \eqref{eq:word-sampling} and \eqref{eq:reward} together, we sample sequentially from
\begin{align}
	\infp(&w_t|w_{1:t-1})\!\propto\notag\\
	&\exp\left(\frac{\alpha\p(w_t|w_{1:t-1})\! +\! \phi(w_t)^\top\phi(w_t^\star)}{\alpha + \beta}\right)\label{eq:infp}
\end{align}
with the complexity $\mathcal O(dV)$ of a standard softmax. Compared to standard VAE training, Algorithm \ref{algorithm} only needs one additional forward pass (with identical complexity) to sample $\tilde w$ form $\infp$.

Equation \eqref{eq:infp} gives a simple interpretation of our proposed training methodology. We locally correct predictions made by the model proportionally to the distance to the ground-truth in the embeddings space. Hence, we consider the ground-truth and the model prediction for exploration.

\section{Related Work}

Our discussion of exposure bias complements recent work that summarizes modern generative models, for example Caccia et al.~\citeyearpar{caccia18} and Lu et al.~\citeyearpar{lu18}. Shortcomings of maximum-likelihood training for sequence generation have often been discussed \cite{ding17, leblondAOL17, ranzato2015sequence}, but without pointing to generalization as the key aspect. An overview of recent deep reinforcement learning methods for conditional generation can be found in \cite{keneshloo18}.

Our proposed approach follows work by Ding et al.~\citeyearpar{ding17} and Tan et al.~\citeyearpar{tan2018} by employing both, policy and reward for exploration. In contrast to them, we do not use $n$-gram based reward.  Compared to RAML \cite{norouziBCJSWS16}, we do not perturb the ground-truth context, but correct the policy predictions. Scheduled sampling \cite{bengioVJS15} and word-dropout~\cite{bowmanVVDJB15} also apply a correction, yet one that only affects the probability of the ground-truth. Chen et al.~\citeyearpar{chenL0ZL017} propose Bridge modules that similarly to Ding et al.~\citeyearpar{ding17} can incorporate arbitrary ground-truth perturbations, yet in an objective motivated by an auxiliary KL-divergence.

Merity et al.~\citeyearpar{merityKS17} have shown that generalization is crucial to language modeling, but their focus is regularizing parameters and activations. Word-embeddings to measure deviations from the ground-truth have also been used by Inan et al.~\citeyearpar{Inan2016TyingWV}, yet under log-likelihood. Concurrently to our work, Li et al.~\citeyearpar{siyao19} employ  embeddings to design reward functions in abstractive summarization.

\section{Experiments}
\label{sec:experiments}

\paragraph{Parametrization}
The policies of all our models and all baselines use the same RNN. We use a 256 dimensional GRU \cite{cho-etal-2014-learning} and 100-dimensional pre-trained word2vec input embeddings. Optimization is preformed by Adam \cite{kingmaB14} with an initial learning rate of 0.001 for all models. For all methods, including scheduled sampling, we do not anneal hyper-parameters such as the keep-probability for the following reasons. First, in an unconditional setting, using \emph{only} the model's prediction is not a promissing setting, so it is unclear what value to anneal to. Second, the continous search-space of schedules makes it sufficiently harder to compare different methods. For the same reason, we do not investigate annealing the KL term or the $\alpha,\beta$-parametrization of the models. We use the inference network parametrization of \cite{bowmanVVDJB15} which employs a diagonal Gaussian for $q$. 

We found the training regime to be very sensitive to the $\alpha,\beta$-parametrization. In particular, it is easy to pick a set of parameters that   does not truly incorporate exploration, but reduces to maximum likelihood training with only ground truth contexts (see also the discussion of Figure \ref{plot:correct} in Section \ref{sec:results}). After performing a grid-search (as done also for RAML) we choose\footnote{The scale of $\alpha$ is relatively small as the log-probabilities in \eqref{eq:infp} have significantly larger magnitude than the inner products, which are in $[0,1]$ due to the normalization.} $\alpha=0.006,\beta=0.067$ for \textsc{Ours}, the method proposed. In addition, we report for an alternative model \textsc{Ours-B} with $\alpha=0.01,\beta=0.07$. 
 
\paragraph{Data}
For our experiments, we use a one million sentences subset of the BooksCorpus \cite{kiros2015skip, zhu2015aligning} with a 90-10 train-test split and a 40K words vocabulary. The corpus size is chosen to challenge the above policy with both scenarios, overfitting and underfitting.

\subsection{Baselines}
As baselines we use a standard \textsc{VAE} and a VAE with \textsc{RAML} decoding that uses identical reward as our method (see Tan et al.\citeyearpar{tan2018} for details on RAML as a special case). Furthermore, we use two regularizations of the standard VAE, scheduled sampling \textsc{SS-p} and word-dropout \textsc{wdrop-p} as proposed by Bowman et al.~\citeyearpar{bowmanVVDJB15}, both with fixed probability $p$ of using the ground-truth.

In addition, we report as special cases with $\z=\mathbf 0$ results for our model (\textsc{Ours-det}), RAML (\textsc{RAML-det}), scheduled sampling (\textsc{SS-p-det}), and the VAE (\textsc{LM}, a language model).

\subsection{Results}
\label{sec:results}
Figure \ref{plot:latent} shows training and test negative sequence log-likelihood evaluated during training and Table \ref{table:latent} shows the best performance obtained. All figures and tables are averaged across three runs.

\begin{table}[ht]
\def\arraystretch{1.1}
\centering
\footnotesize
\begin{tabular}{lll}
Model & Train NLL & Test NLL\\
\hline
\textsc{Ours} & 48.52  & \textbf{52.54}\\
\textsc{Ours-B} & 49.51  & 52.61\\
\textsc{Ours-det} & 48.06  & 52.87\\
\textsc{SS-0.99} & 48.11 & 52.60\\
\textsc{SS-0.98} & 48.21 & 52.62\\
\textsc{SS-0.95} & 48.38 & 52.69\\
\textsc{SS-0.90} & 49.02 & 52.89\\
\textsc{SS-0.99-det} & 48.08 & 52.90\\
\textsc{RAML} & 48.26 & 52.56\\
\textsc{RAML-det} & 48.26 & 52.86\\
\textsc{wdrop-0.99} & 48.19 & 52.86\\
\textsc{LM} & \textbf{47.65} & 53.01\\
\textsc{VAE} & 47.86 & 52.66\\
\textsc{wdrop-0.9} & 50.86 & 54.65\\
\end{tabular}
\caption{Training and test performance}
\label{table:latent}
\end{table}

We observe that all latent models outperform their deterministic counterparts (crossed curves) in  terms of both, generalization and overall test performance. This is not surprising as regularization is one of the benefits of modeling uncertainty through latent variables. Scheduled sampling does improve generalization for $p\approx 1$ with diminishing returns at $p=0.95$ and in general performed better than word dropout. Our proposed models outperform all others in terms of generalization and test performance. Note that the performance difference over RAML, the second best method, is solely due to incorporating also model-predicted contexts during training.

Despite some slightly improved performance, all latent models except for \textsc{Ours-B} have a KL-term relatively close to zero. \textsc{Ours-B} is $\alpha$-$\beta$-parametrized to incorporte slightly more model predictions at higher temperatur and manages to achieve a KL-term of about 1 to 1.5 bits.  These findings are similar to what \cite{bowmanVVDJB15} report \emph{with} annealing but still significantly behind work that addresses this specific problem \cite{YangHSB17,shen2018}. Appendix \ref{appendix-kl} illustrates how our models can obtain larger KL-terms -- yet at degraded performance -- by controlling exploration. We conclude that improved autoregressive modeling inside the ERPO framework cannot alone overcome VAE-collapse.

\begin{figure}[ht]
\centering
\begin{tikzpicture}

\begin{axis}[
    xlabel={\small Training Time},
    height = 3.5cm,
    width=8.5	cm,
    ylabel style={yshift=-0.2cm},
    xlabel style={yshift=0.2cm},
    legend cell align={left},
    legend style={font=\scriptsize\sc, row sep=-2.4pt},
     smooth, each nth point=4,
    xmin=20000,
    xmax=180000,
    xticklabels={,,},
    restrict x to domain=20000:180000,
    scaled x ticks=base 10:0,
    xticklabel style={/pgf/number format/fixed},
    ylabel style={yshift=-0.2cm},
    xlabel style={yshift=0.2cm},
     legend columns=4, 
        legend style={
        mark size=1pt,
        line width=.1pt,
        mark options={scale=0.1} ,
        at={($(0,0)+(3.3cm,2.05cm)$)},
        ,anchor=south,align=center,
        },
]

\pgfplotstableread{correct.data}{\datatable}
\addplot [giallo, thick] table [x={x}, y={latent-alpha0.01-beta0.1-gamma1.5}] {\datatable};
\addlegendentry{Ours}
\addplot [rosso, thick] table [x={x}, y={latent-alpha0.015-beta0.1-gamma1.4}] {\datatable};
\addlegendentry{Ours-B}
\addplot [viola, thick] table [x={x}, y={lm-alpha0.1-beta0.05-gamma1.8}] {\datatable};
\addlegendentry{Ours-det}
\addplot [verde] table [x={x}, y={latent-raml-beta0.065}] {\datatable};
\addlegendentry{RAML}           
\addplot [babyblue, dashed] table [x={x}, y={latent-ss-p0.98}] {\datatable};
\addlegendentry{SS-0.98}
\addplot [babyblue] table [x={x}, y={latent-ss-p0.99}] {\datatable};
\addlegendentry{SS-0.99}
\addplot [babyblue, dashed] table [x={x}, y={latent-ss-p0.95}] {\datatable};
\addlegendentry{SS-0.95}
\addplot [black] table [x={x}, y={lm}] {\datatable};
\addlegendentry{LM/VAE}

\end{axis}

\end{tikzpicture}
\caption{Fraction of correct words during training. Numbers include \textit{forced} and \textit{correctly predicted} words.}
\label{plot:correct}
\end{figure}
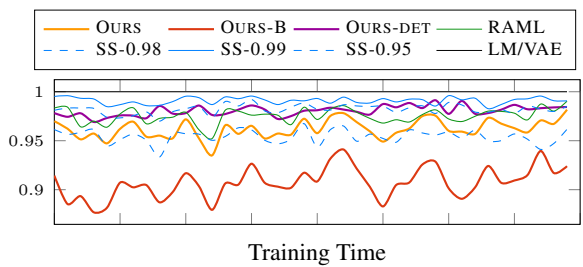

\normalsize We have discussed many approaches that deviate from training exclusively on ground-truth contexts. Therefore, an interesting quantity to monitor across methods is the fraction of words that correspond to the ground-truth. Figure \ref{plot:correct} shows these fractions during training for the configurations that gave the best results. Interestingly, in the latent setting our method relies by far the least on ground-truth contexts whereas in the deterministic setting the difference is small.

\section{Conclusion}
We have argued that exposure bias does not point to a problem with the standard methodology of training autoregressive sequence model. Instead, it refers to a notion of generalization to unseen sequences that does not manifest in log-likelihood training and testing, yet might be desirable in order to capture the flexibility of natural language.

To rigorously incorporate the desired generalization behavior, we have proposed to follow the reinforcement learning formulation of Tan et al.~\citeyearpar{tan2018}. Combined with an embedding-based reward function, we have shown excellent generalization performance compared to the unregularized model and better generalization than existing techniques on language modeling and sentence autoencoding.

\paragraph{Future work}
We have shown that the simple reward function proposed here leads to a form of regularization that fosters generalization when evaluated inside the maximum-likelihood framework. In the future, we hope to conduct a human evaluation to assess the generalization capabilities of models trained under maximum-likelihood and reinforcement learning more rigorously. Only such a framework-independent evaluation can reveal the true gains of carefully designing reward functions compared to simply performing maximum-likelihood training.

\bibliographystyle{acl_natbib}
\bibliography{bibliography}

\phantom{x}\newpage\phantom{x}\newpage\newpage\appendix
\section{KL-term Under Broader Exploration}
\label{appendix-kl}
As mentioned in Section \ref{sec:experiments}, our model \textsc{Ours-B} is the sole model to achieve a significant KL-term, yet far from the amount of information one would hope to encode. However, larger KL-terms can be obtained at degraded performance as similarly reported by \cite{bowmanVVDJB15}. To illustrate this, we fix a temperature $\beta=0.065$ and vary $\beta$ between 0.001 and 10.0. Since $\alpha$ controls the degree of exploration (amount of model generated contexts), the information encoded in the contexts can be varied and the code becomes more important. Similar plots can be obtained for the other methods.

\pgfplotsset{width=7cm,compat=1.3}
\begin{figure}[ht]
\centering
\hspace{-1cm}

\begin{tikzpicture}

\begin{axis}[
  axis y line*=left,
  ymin=52, ymax=70,
  xlabel=$\alpha$,
  ylabel=NLL,
  xmode=log,
  ymode=log,
  ylabel style={font=\small,yshift=-0.7cm},
  yticklabel style={/pgf/number format/fixed},
  scaled ticks=false,
  width=7.7cm
]
\addplot[mark=x,rosso]
  coordinates{
(0.001, 53.24)
(0.01, 53.01)
(0.1, 57.82)
(0.15, 61.83)
(0.5, 67.26)
(1.0, 69.18)
(10.0, 69.6)
}; \label{plot_one}

\end{axis}

\begin{axis}[
  axis y line*=right,
  axis x line=none,
  ymin=0, ymax=8,
  ylabel=bits,
  xmode=log,
  ymode=log,
  legend style={at={(1,0)},anchor=south east, font=\scriptsize},
  ylabel style={font=\small,yshift=0.3cm},
  yticklabel style={/pgf/number format/fixed},
  scaled ticks=false,
  width=7.7cm
]
\addplot[mark=o,babyblue]
  coordinates{
(0.001, 0)
(0.01, 0)
(0.1, 3.38)
(0.15, 4.42)
(0.5, 7.62)
(1.0, 6.8)
(10.0, 7.16)
}; \label{plot_two}
\addlegendentry{KL-term}
\addlegendimage{/pgfplots/refstyle=plot_one}\addlegendentry{Test NLL}
\end{axis}

\end{tikzpicture}
\caption{Test NLL and KL-term for various values of $\alpha$ on $\log$-scale. }
\label{fig:kl-term}
\end{figure}
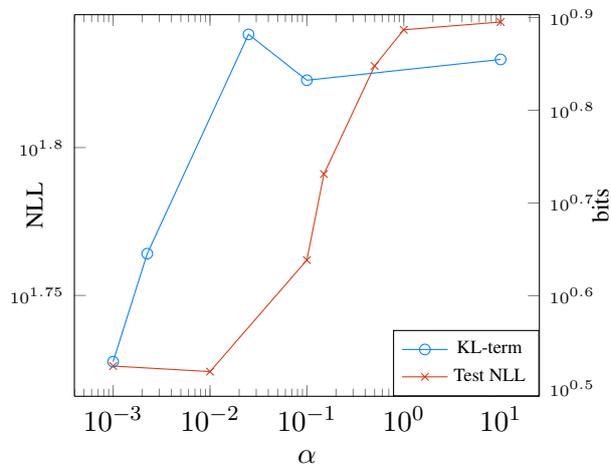

\end{document}